\pdfoutput=1

\documentclass[11pt,a4paper]{article}

\usepackage{acl}

\usepackage{times}
\usepackage{latexsym}

\usepackage[T1]{fontenc}

\usepackage[utf8]{inputenc}

\usepackage{caption}
\usepackage{subcaption}

\usepackage{microtype}

\usepackage{todonotes}
\usepackage{glossaries}

\usepackage{amssymb}
\usepackage{amsmath}
\usepackage{bm}
\usepackage{booktabs} 
\usepackage{multirow}
\usepackage{makecell}
\usepackage{pmboxdraw}

\newacronym{nmt}{NMT}{Neural Machine Translation}
\newacronym{bpe}{BPE}{Byte Pair Encoding}
\newacronym{nvs}{NVS}{Neural Vocabulary Selection}
\newacronym{bow}{BOW}{Bag-of-words}
\newacronym{kd}{KD}{Knowledge Distillation}
\newacronym{ssru}{SSRU}{Simpler Simple Recurrent Unit}
\newacronym{aan}{AAN}{Average Attention Network}

\usepackage{xurl}
\usepackage{hyperref}
\usepackage{natbib}
\title{The Devil is in the Details:\\On the Pitfalls of Vocabulary Selection in Neural Machine Translation}

 \author{{\bf Tobias Domhan\thanks{~~Equal contributions.} \hspace{1mm}} {\bf Eva Hasler\footnotemark[1] \hspace{1mm}} {\bf Ke Tran \hspace{1mm}} {\bf Sony Trenous \hspace{1mm}}  {\bf Bill Byrne \hspace{1mm}} {\bf Felix Hieber} \\
         Amazon AI Translate \\
Berlin, Germany \\
 \texttt{\{domhant,ehasler,trnke,trenous,willbyrn,fhieber\}@amazon.com}}

\begin{document}
\maketitle
\begin{abstract}
Vocabulary selection, or lexical shortlisting, is a well-known technique to improve latency of Neural Machine Translation models by constraining the set of allowed output words during inference. The chosen set is typically determined by separately trained alignment model parameters, independent of the source-sentence context at inference time. While vocabulary selection appears competitive with respect to automatic quality metrics in prior work, we show that it can fail to select the right set of output words, particularly for semantically non-compositional linguistic phenomena such as idiomatic expressions, leading to reduced translation quality as perceived by humans. Trading off latency for quality by increasing the size of the allowed set is often not an option in real-world scenarios. We propose a model of vocabulary selection, integrated into the neural translation model, that predicts the set of allowed output words from contextualized encoder representations. This restores translation quality of an unconstrained system, as measured by human evaluations on WMT newstest2020 and idiomatic expressions, at an inference latency competitive with alignment-based selection using aggressive thresholds, thereby removing the dependency on separately trained alignment models.
\end{abstract}

\section{Introduction}
\gls*{nmt} has achieved great improvements in translation quality, largely thanks to the introduction of Transformer models \cite{Vaswani:2017}.
However, increasingly larger models \cite{aharoni-etal-2019-massively,Arivazhagan2019MassivelyMN} lead to prohibitively slow inference when deployed in industrial settings.
Especially for real-time applications, low latency is key.
A number of inference optimization speed-ups have been proposed and are used in practice: reduced precision \cite{aji-heafield-2020-compressing}, replacing self-attention with \glspl*{aan}~\cite{zhang-etal-2018-accelerating}, \glspl*{ssru}~\cite{kim-etal-2019-research}, or model pruning \cite{behnke-heafield-2020-losing, behnke2021efficient}.

\begin{figure}[t]
\centering
\begin{tabular}{l}
EN: to swal\pmboxdrawuni{2581} low the bitter pill \\
DE: in den sau\pmboxdrawuni{2581} ren \textbf{\color{orange}ap\pmboxdrawuni{2581}} \textit{\color{red}fel} \textbf{\color{orange}bei\pmboxdrawuni{2581}} ßen \\
GL: to \textbf{\color{orange}bi}te into the sour \textbf{\color{orange}ap} \textit{\color{red}ple}\\\midrule
EN: by ho\pmboxdrawuni{2581} ok or cro\pmboxdrawuni{2581} ok \\
DE: auf \textbf{\color{orange}biegen} und \textit{\color{red}brechen} \\
GL: by \textbf{\color{orange}bending} and \textit{\color{red}breaking}\\\midrule
EN: to buy a p\pmboxdrawuni{2581} ig in a po\pmboxdrawuni{2581} ke \\
DE: die \textbf{\color{orange}kat\pmboxdrawuni{2581}} \textbf{\color{orange}ze} im \textit{\color{red}sack} kaufen \\
GL: to buy the \textbf{\color{orange}cat} in the \textit{\color{red}bag} \\\midrule
EN: to swe\pmboxdrawuni{2581} at blood\\
DE: blu\pmboxdrawuni{2581} t und \textbf{\color{orange}wasser} sch\pmboxdrawuni{2581} wit\pmboxdrawuni{2581} zen \\
GL: to sweat blood and \textbf{\color{orange}water} \\\midrule
EN: make yourself at home !\\
DE: machen sie es sich \textbf{\color{orange}bequem} ! \\
GL: make yourself \textbf{\color{orange}comfortable} !
\end{tabular}
\caption{Examples of subword-segmented idiomatic expressions (EN) and their German correspondences (DE) as well as an English gloss (GL) of the German expression. Alignment-based vocabulary selection: output tokens missing from the allowed set of top-k output tokens are marked in orange/bold (red/italic) for k=200 (k=1000).}
\label{fig:coverage_examples}
\end{figure}

Another technique that is very common in practice is vocabulary selection \cite{jean-etal-2015-using} which usually provides a good tradeoff between latency and automatic metric scores (BLEU) and reduced inference cost is often preferred over the loss of ${\sim}0.1$ BLEU.
Vocabulary selection is effective because latency is dominated by expensive, repeated decoder steps, where the final projection to the output vocabulary size contributes to a large portion of time spent~\citep{berard2021_efficient}. Despite high parallelization in GPUs, vocabulary selection is still relevant for GPU inference for state-of-the-art models.

However, we show that standard methods of vocabulary selection based on alignment model dictionaries lead to quality degradations not sufficiently captured by automatic metrics such as BLEU. We demonstrate that this is particularly true for semantically non-compositional linguistic phenomena such as idiomatic expressions, and aggressive thresholds for vocabulary selection. For example, see Figure \ref{fig:coverage_examples} for alignment-model based vocabulary selection failing to include tokens crucial for translating idiomatic expressions in the set of allowed output words. While less aggressive thresholds can reduce the observed quality issues, it also reduces the desired latency benefit. In this paper we propose a neural vocabulary selection model that is jointly trained with the translation model and achieves translation quality at the level of an unconstrained baseline with latency at the level of an aggressively thresholded alignment-based vocabulary selection model.

Our contributions are as follows:
\begin{itemize}
  \item We demonstrate that alignment-based vocabulary selection is not limited by alignment model quality, but rather inherently by making target word predictions out of context~(\S\ref{sec:pitfalls}).
  \item We propose a Neural Vocabulary Selection (NVS) model based on the contextualized deep encoder representation~(\S\ref{sec:model}).
  \item We show that alignment-based vocabulary selection leads to human-perceived translation quality drops not sufficiently captured by automatic metrics and that our proposed model can match an unconstrained model's quality while keeping the latency benefits of vocabulary selection~(\S\ref{sec:experiments}).
\end{itemize}

\section{Pitfalls of vocabulary selection}
\label{sec:pitfalls}
We first describe vocabulary selection and then analyze its shortcomings.
Throughout the paper, we use the recall of unique target sentence tokens as a proxy for measuring vocabulary selection quality, i.e. the reachability of the optimal translation.
We use the average vocabulary size in inference decoder steps across sentences as a proxy for translation latency since it directly impacts decoding speed~\cite{kasai2020deep}.

\subsection{Vocabulary selection}
Vocabulary selection \cite{jean-etal-2015-using}, also known as lexical shortlisting or candidate selection, is a common technique for speeding up inference in sequence-to-sequence models, where the repeated computation of the softmax over the output vocabulary $\mathcal{V}$ of size $V$ incurs high computational cost in the next word prediction at inference time:
\begin{equation}
    p(y_t|y_{1:t-1}, x; \theta) = \text{softmax}(\bm{W}\bm{h} + \bm{b}),
\label{eq:softmax}
\end{equation}
where $\bm{W} \in \mathbb{R}^{V \times d}$, $\bm{b} \in \mathbb{R}^V$ and $\bm{h} \in \mathbb{R}^d$, $d$ being the hidden size of the network.
Vocabulary selection chooses a subset $\bar{\mathcal{V}}\subset \mathcal{V}$, with $\bar{V}\ll V$, to reduce the size of matrix multiplication in Equation \eqref{eq:softmax} such that 
\begin{equation}
    \bar{p}(y_t|y_{1:t-1}, x; \theta) = \text{softmax}(\bm{\bar{W}}\bm{h} + \bm{\bar{b}}),
\label{eq:reduced_softmax}
\end{equation}
where $\bm{\bar{W}}\in\mathbb{R}^{\bar{V}\times d}$ and $\bar{b}\in\mathbb{R}^{\bar{V}}$.
The subset $\bar{\mathcal{V}}$ is typically chosen to be the union of the top-k target word translations for each source token, according to the word translation probabilities of a separately trained word alignment model~\citep{jean-etal-2015-using, shi-knight-2017-speeding}.
Decoding with vocabulary selection usually yields similar scores according to automatic metrics, such as BLEU \cite{papineni2002bleu}, compared to unrestricted decoding but at reduced latency~\cite{vocabularySelectionStrategies, mi-etal-2016-vocabulary, sankaran2017attention, junczys-dowmunt-etal-2018-marian-cost}. 
In the following, we show that despite its generally solid performance, vocabulary selection based on word alignment models negatively affects translation quality, not captured by standard automatic metrics.
We use models trained on WMT20~\cite{barrault-etal-2020-findings} data for all evaluations in this section, see Section~\ref{exp_setup} for details. 

\subsection{Alignment model quality}

\begin{table}
\small
\centering
\begin{tabular}{llrrrr}
& &  \multicolumn{2}{c}{$k = 200$}  & \multicolumn{2}{c}{$k = 1000$} \\
& \textbf{Scope}  &\textbf{model} & \textbf{ref.} & \textbf{model} & \textbf{ref.}   \\
\midrule
\multirow{3}{*}{EN-DE} & fast\_align & 99.6 & 97.5  & 99.9 & 99.7 \\
& GIZA++ & 99.6 & 97.8 &  100.0 & 99.7 \\
& MaskAlign & 96.9 & 93.1  & 99.6 & 98.6 \\
\midrule
\multirow{3}{*}{EN-RU} & fast\_align & 98.7 &  93.8 & 99.9 &  98.9 \\
& GIZA++ & 98.6 & 94.2  &  99.9 & 99.2 \\
& MaskAlign & 94.2 & 87.2  &  99.1 & 96.7 \\
\bottomrule
\end{tabular}
\caption{Recall of unique reference and model tokens on newstest2020 with word probability tables extracted from different alignment models.}
\label{tab:alignment-model-quality}
\end{table}

In practice, the chosen subset of allowed output words is often determined by an alignment model, such as \texttt{fast\_align}~\citep{dyer-etal-2013-simple}, which provides a trade-off between the speed of alignment model training and the quality of alignments~\cite{jean-etal-2015-using,junczys-dowmunt-etal-2018-marian-cost}.
\texttt{fast\_align}'s reparametrization of IBM model 2~\citep{brown1993mathematics} places a strong prior for alignments along the diagonal.
We investigate whether more sophisticated alignment models can lead to better vocabulary selection, especially for language pairs with high amount of reordering.
To evaluate this we compute the recall of translation model and reference tokens using GIZA++~\citep{och03:asc} and MaskAlign\footnote{With default hyper-parameters from \url{https://github.com/THUNLP-MT/Mask-Align}}~\cite{chen-etal-2021-mask} as seen in Table~\ref{tab:alignment-model-quality}.
We extract top-k word translation tables (from \texttt{fast\_align}, GIZA++, and MaskAlign) by force-aligning the training data.
Overall, GIZA++ achieves the best recall, and it is just slightly better than \texttt{fast\_align}. MaskAlign, a state-of-the-art neural alignment model, underperforms \texttt{fast\_align} with respect to recall.
While performance of MaskAlign may be improved with careful tuning of its hyperparameters via gold alignments~\citep{chen-etal-2021-mask}, we choose \texttt{fast\_align} as a strong, simple baseline for vocabulary selection in the following.

\subsection{Out-of-context word selection}
Alignment-based vocabulary selection does not take source sentence context into account. A top-k list of translation candidates for a source word will likely cover multiple senses for common words, but may be too limited when a translation is highly dependent on the source context. Here we consider \textit{idiomatic expressions} as a linguistic phenomenon that is highly context-dependent due to its semantically non-compositional nature. 

Table~\ref{tab:idioms_recall_1} compares the recall of tokens in the reference translation when querying the translation lexicon of the alignment model for two different top-k settings. Recall is computed as the percentage of unique tokens in the reference translation that appear in the top-k lexicon, or more generally, in the set of predicted tokens according to a vocabulary selection model. We evaluate two scopes for test sets of idiomatic expressions: the full source and target sentence vs. the source and target idiomatic multi-word expressions according to metadata.
The Idioms test set is an internal set of 100 English idioms in context and their human translations. ITDS is the IdiomTranslationDS\footnote{\url{https://github.com/marziehf/IdiomTranslationDS}} data released by~\citet{fadaee-etal-2018-examining} with 1500 test sentences containing English and German idiomatic expressions for evaluation into and out of German, respectively.
The results show that recall increases when increasing \textit{k} but is consistently lower for the idiomatic expressions than for full sentences. Clearly, the idiom translations contain tokens that are on average less common than the translations of ``regular'' inputs.
As a consequence, increasing the output vocabulary is less effective for idiom translations, with recall lagging behind by up to 9.3\%. This can directly affect translation quality because the \gls*{nmt} model will not be able to produce idiomatic translations given an overly restrictive output vocabulary.

\begin{table}
\small
\centering
\begin{tabular}{@{} l lc c c c @{}}
& &  \multicolumn{2}{c}{$k = 200$}  & \multicolumn{2}{c}{$k = 1000$} \\
& \textbf{Scope}  &\textbf{Idioms} & \textbf{ITDS} & \textbf{Idioms} & \textbf{ITDS}   \\
\midrule
\multirow{2}{*}{EN-DE} & sentence &  96.0 & 91.5 & 99.1 & 97.3 \\
& idiom & 80.0 & 58.2  & 92.8 & 88.0 \\
\midrule
\multirow{2}{*}{DE-EN} & sentence & - & 92.2 & - & 98.0 \\
& idiom & - & 75.5 & - & 90.1 \\
\midrule
\multirow{2}{*}{EN-RU} & sentence & 93.0 & - & 98.7 & - \\
& idiom & 65.7 & -  & 85.6 & - \\
\bottomrule
\end{tabular}
\caption{Recall on internal test set of idioms in context and IdiomTranslationDS test set.}
\label{tab:idioms_recall_1}
\end{table}

Table~\ref{tab:idioms_recall_2} shows a similar comparison but here we evaluate full literal translations vs. full idiomatic translations on a data set of English proverbs from Wikiquote\footnote{\url{https://github.com/awslabs/sockeye/tree/naacl2022/naacl2022/wikiquote}}. For EN-DE, we extracted 94 triples of English sentence and two references, for EN-RU we extracted 262 triples. Although in both cases recall can be improved by increasing \textit{k}, it helps considerably less for idiomatic than for literal translations.

\begin{table}
\small
\centering
\begin{tabular}{l l c c }
& &  \multicolumn{2}{c}{\bf Wikiquote}  \\
 & \textbf{Scope}  & $k=200$ & $k=1000$   \\
\midrule
\multirow{2}{*}{EN-DE} & literal &  96.5 & 99.2 \\
& idiomatic &  74.6 & 91.0 \\
\midrule
\multirow{2}{*}{EN-RU} & literal & 91.2 &  99.2 \\
& idiomatic & 67.6 & 88.9 \\
\bottomrule
\end{tabular}
\caption{Recall on English proverbs and their literal and idiomatic translations sourced from Wikiquote.}
\label{tab:idioms_recall_2}
\end{table}

Figure~\ref{fig:coverage_examples} shows examples of idiomatic expressions from the ITDS set and the output tokens belonging to an idiomatic translation that are missing from the respective lexicon used for vocabulary selection. While for some of the examples, increasing the lexicon size solves the problem, for others the idiomatic translation can still not be generated because of missing output tokens.

These results demonstrate that there is room for improvement in vocabulary selection approaches when it comes to non-literal translations.

\subsection{Domain mismatch in adaptation settings}
Using a word alignment model to constrain the \gls*{nmt} output vocabulary means that this model should ideally also be adapted when adapting the \gls*{nmt} model to a new domain. Table~\ref{tab:idioms_recall_3} shows that adapting the word alignment model with relevant in-domain data (in this case, idiomatic expressions in context) yields strong recall improvements for vocabulary selection. Compared to increasing the per-source-word vocabulary as shown in Table~\ref{tab:idioms_recall_1}, the improvement in recall for idiom tokens is larger which highlights the importance of having a vocabulary selection model which matches the domain of the NMT model. This also corroborates the finding of \citet{bogoychev-chen-2021-highs} that vocabulary selection can be harmful in domain-mismatched scenarios.

We argue that integrating vocabulary prediction into the \gls*{nmt} model avoids the need for mitigating domain mismatch because domain adaptation will update both parts of the model. This simplifies domain adaptation since it only needs to be done once for a single model and does not require adaptation or re-training of a separate alignment model.

\begin{table}
\small
\centering
\begin{tabular}{l l c c }
& &  \multicolumn{2}{c}{\bf idioms in context}  \\
 & \textbf{Scope}  & w\slash o adapt  & w\slash~adapt   \\
\midrule
\multirow{2}{*}{EN-DE} & sentence &  96.0 &  98.0 \\
& idiom &  80.0  & 97.2\\
\midrule
\multirow{2}{*}{EN-RU} & sentence & 93.0& 96.6\\
& idiom &  65.7 & 91.7\\
\bottomrule
\end{tabular}
\caption{Recall on internal test set of idioms in context, with and without adapting the \texttt{fast\_align} translation lexicons ($k=200$).}
\label{tab:idioms_recall_3}
\end{table}

\subsection{Summary}
We use target recall as a measure for selection model quality.
We see that alignment model quality only has a limited impact on target token recall with more recent models actually having lower recall overall.
In domain adaptation scenarios vocabulary selection limits translation quality if the selection model is not adapted.
The main challenge for alignment-based vocabulary selection comes from its out-of-context selection of target tokens on a token-by-token basis, shown to reduce recall for translation of idiomatic, non-literal expressions. Increasing the size of the allowed set can compensate for this shortcoming at the cost of latency.
However, this begs the question of whether context-sensitive selection of target tokens can achieve higher recall without increasing vocabulary size.

\section{Neural Vocabulary Selection (NVS)}
\label{sec:model}
We incorporate vocabulary selection directly into the neural translation model, instead of relying on a separate statistical model based on token translation probabilities.
This enables predictions based on contextualized representations of the full source sentence.
It further simplifies the training procedure and domain adaptation, as we do not require a separate training procedure for an alignment model.

\begin{figure*}[th] 
\centering
  \includegraphics[width=0.9\linewidth]{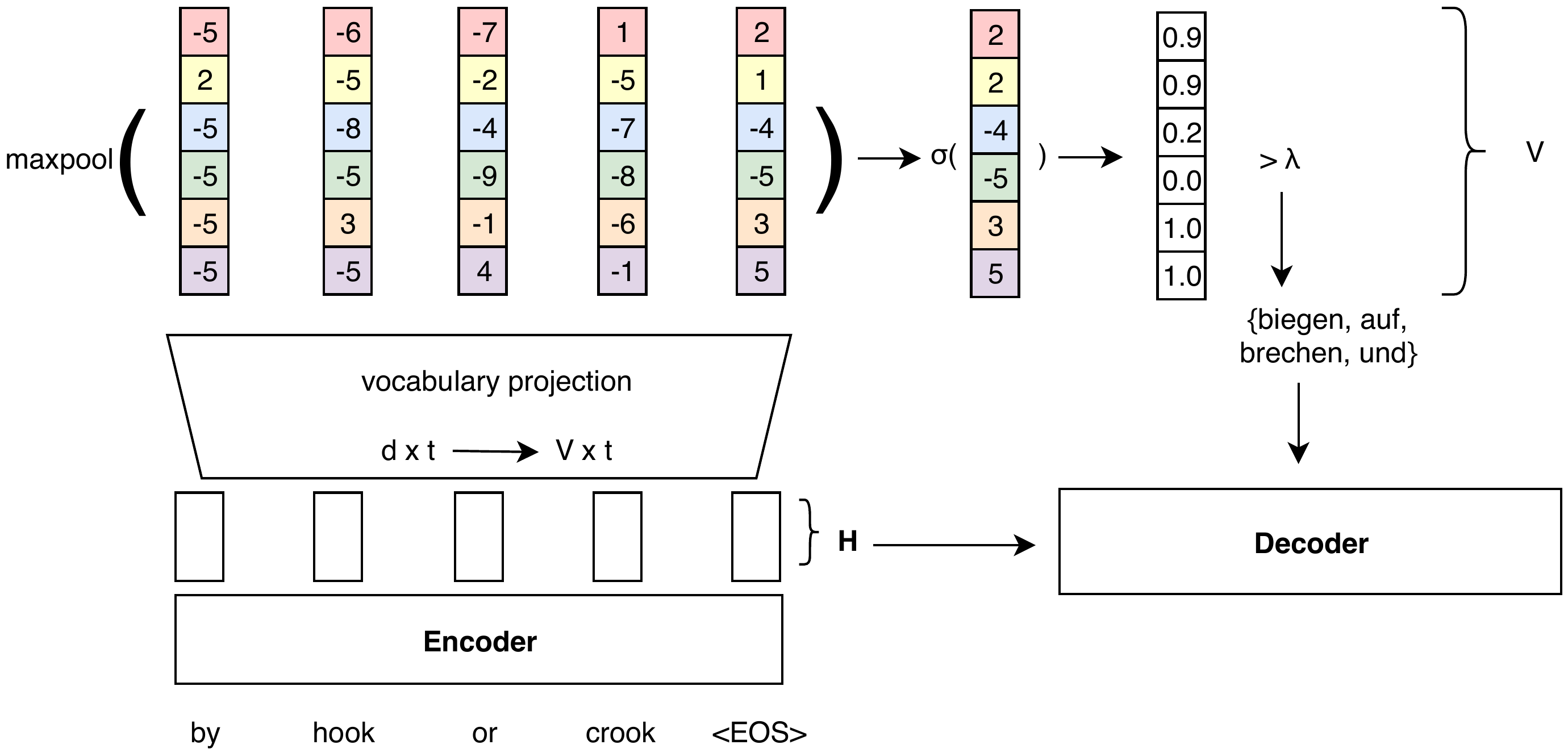}
  \caption{The proposed Neural Vocabulary Selection (NVS) model. A subset of the full vocabulary is selected based on encoder representation $H$ and passed to the decoder for a reduced output projection at every decoder step.}
  \label{fig:model}
\end{figure*}

The goal of our approach is three-fold. We aim to (1) keep the general Transformer~\citep{Vaswani:2017} translation model architecture,
    (2) incur only a minimal latency overhead that amortizes by cheaper decoder steps due to smaller output vocabularies, and
    (3) scale well to sentences of different lengths.

Figure~\ref{fig:model} shows the \gls*{nvs} model. We base the prediction of output tokens on the contextualized hidden representation produced by the Transformer encoder $\bm{H} \in \mathbb{R}^{d \times t}$ for $t$ source tokens and a hidden size of $d$.
The $t$ source tokens are comprised of $t-1$ input tokens and a special \texttt{<EOS>} token.
To obtain the set of target tokens, we first project each source position to the target vocabulary size $V$, apply max-pooling across tokens~\cite{shen2018baseline}, and finally use the sigmoid function, $\sigma (\cdot)$, to obtain
\begin{equation}
    \bm{z} = \sigma (\text{maxpool}(\bm{W} \bm{H} + \bm{b})),
\end{equation}
where $\bm{W} \in \mathbb{R}^{V \times d}$, $\bm{b} \in \mathbb{R}^V$ and $\bm{z} \in \mathbb{R}^V$.
The max-pooling operation takes the per-dimension maximum across the source tokens, going from $\mathbb{R}^{V \times t}$ to $\mathbb{R}^{V}$.
Each dimension of $\bm{z}$ indicates the probability of a given target token being present in the output given the source.
To obtain the target \gls*{bow}, we select all tokens where $\bm{z}_i > \lambda$ as indicated in the right-hand side of Figure~\ref{fig:model}, where $\lambda$ is a free parameter that controls the size $\bar{V}$ of the reduced vocabulary $\bar{\mathcal{V}}$. At inference time, the output projection and softmax at every decoder step are computed over the predicted \gls*{bow} of size $\bar{V}$ only.

We achieve goal (1) by basing predictions on the encoder representation already used by the decoder.
Goal (2) is accomplished by restricting \gls*{nvs} to a single layer and basing the prediction on the encoder output, where we can parallelize computation across source tokens.
Inference latency is dominated by non-parallelizable decoder steps~\cite{kasai2020deep}.
By projecting to the target vocabulary per source token, each source token can ``vote'' on a set of target tokens. The model automatically scales to longer sentences via the max-pooling operation, acting as a union of per-token choices, fulfilling goal (3).
Max-pooling does not tie the predictions across timesteps as they would be with mean-pooling which would also depend on sentence length.
Additionally, we factor in a sentence-level target token prediction based on the \texttt{<EOS>} token.
The probability of a target word being present is represented by the source position with the highest evidence, backing off to a base probability of a given word via the bias vector $\bm{b}$.

To learn the $V\times d + V$ parameters for \gls*{nvs}, we use a binary cross-entropy loss with the binary ground truth vector $\bm{y} \in \mathbb{R}^V$, where each entry indicates the presence or absence of target token $\bm{y}_i$.
We define the loss as
\begin{equation*}\label{eq:loss_nvs}
    L_{\text{\gls*{nvs}}} = \frac{1}{Z} \sum_{i=0}^V \bm{y}_i \log(\bm{z}_i) \lambda_p + (1 - \bm{y}_i) \log(1 - \bm{z}_i),
\end{equation*}
where $\lambda_p$ is a weight for the positive class and $Z = V + (\lambda_p-1) * n_p$ is the normalizing factor, with $n_p = \sum_i \bm{y}_i$ being the number of unique target words.
Most dimensions of the $V$-dimensional vector $\bm{y}$ will be zero as only a small number of target words are present for a given sentence.
To counter the class imbalance of the negative class over the positive class the $\lambda_p$ weight allows overweighting the positive class.
This has the same effect as if each target word had occurred $\lambda_p$ times.
The \gls*{nvs} objective ($L_{\text{\gls*{nvs}}}$) is optimized jointly with the standard negative log-likelihood translation model loss ($L_{\text{MT}}$): $L = L_{\text{\gls*{nvs}}} + L_{\text{MT}}.$

\begin{table*}[th!]
\centering
\small
\setlength\tabcolsep{2.7pt} %
 \begin{tabular}{@{} c l | c c c c | c c | c c c @{}}

                                            &                               & \multicolumn{5}{c}{\textbf{newstest2020}}              &                                                        & \multicolumn{3}{c}{\textbf{ITDS idiom test}} \\ 
                                            & \thead{\textbf{vocabulary}\\\textbf{selection}} & \textbf{BLEU} $\uparrow$                               & \textbf{COMET} $\uparrow$                                       & \multicolumn{2}{c}{\textbf{human eval} (\%) $\uparrow$ }                   & \thead{\textbf{CPU}  $\downarrow$ \\ (ms / \%)}                        & \thead{\textbf{GPU}  $\downarrow$ \\ (ms / \%)}                                    & \textbf{BLEU} $\uparrow$                                                       & \multicolumn{2}{c}{\textbf{human eval} (\%) $\uparrow$ }  \\
    \midrule
           
     \multirow{6}{*}{\rotatebox{90}{EN-DE}} & -                             & $34.4${\scriptsize$\color{lightgray} \pm0.2$} & $0.461${\scriptsize $\color{lightgray} \pm 0.003$}     & $\color{lightgray} 100$                                           & $\color{lightgray} 100$                                       & $\color{lightgray} 999 ${\scriptsize $\color{lightgray}   \pm 12$   }     & $ \color{lightgray} 208 ${\scriptsize $\color{lightgray}   \pm 1 $  } & $25.5${\scriptsize $ \color{lightgray} \pm 0.2$}   & $\color{lightgray}100$                                       & $\color{lightgray} 100$   \\ 
                                            & align k=200                   & $34.2${\scriptsize$\color{lightgray} \pm0.1$} & $0.453${\scriptsize $\color{lightgray} \pm 0.004$  }   & $\phantom{0}98.0${\scriptsize $ \color{lightgray} \pm 1.9$  }     & $\phantom{0}96.7${\scriptsize $ \color{lightgray} \pm 1.9$  } & $51.0${\scriptsize $  \color{lightgray} \pm 0.6$ }                        & $ 78.2${\scriptsize $ \color{lightgray} \pm 0.9 $  }                  & $25.4${\scriptsize $ \color{lightgray} \pm 0.2$}   & $\phantom{0}96.6 ${\scriptsize $ \color{lightgray} \pm 1.9$} & $\phantom{0}97.2 ${\scriptsize $ \color{lightgray} \pm 1.8$} \\
                                            & align k=1000                  & $34.3${\scriptsize$\color{lightgray} \pm0.2$} & $0.462${\scriptsize $\color{lightgray} \pm 0.005$  }   & $100.1${\scriptsize $ \color{lightgray} \pm 2.0$    }             & $100.0${\scriptsize $ \color{lightgray} \pm 1.9$  }           & $66.6${\scriptsize $  \color{lightgray} \pm 0.5$ }                        & $ 82.9${\scriptsize $ \color{lightgray} \pm 0.7$   }                  & $25.5${\scriptsize $ \color{lightgray} \pm 0.1$}   & $\phantom{0}98.5 ${\scriptsize $\color{lightgray} \pm 1.7$ } &  $100.4 ${\scriptsize $ \color{lightgray} \pm 1.9$} \\
                                            & NVS $\lambda$=0.99            & $34.4${\scriptsize$\color{lightgray} \pm0.1$} & $0.460${\scriptsize $\color{lightgray} \pm 0.004$  }   & -                                                                 & $\phantom{0}99.5${\scriptsize $ \color{lightgray} \pm 1.7$  } & $46.6${\scriptsize $  \color{lightgray} \pm 0.5$ }                        & $ 79.8${\scriptsize $ \color{lightgray} \pm 0.9  $ }                  & $25.5${\scriptsize $ \color{lightgray} \pm 0.2$}   & -                                                            & \phantom{0}$99.3${\scriptsize $ \color{lightgray} \pm 1.9$          } \\
                                            & NVS $\lambda$=0.9             & $34.4${\scriptsize$\color{lightgray} \pm0.2$} & $0.461${\scriptsize $\color{lightgray} \pm 0.004 $ }   & $100.0 ${\scriptsize $\color{lightgray} \pm 1.8$   }              & -                                                             & $54.1${\scriptsize $  \color{lightgray} \pm 0.7$ }                        & $ 82.2${\scriptsize $ \color{lightgray} \pm 0.9 $  }                  & $25.5${\scriptsize $ \color{lightgray} \pm 0.2$}   & $\phantom{0}99.1${\scriptsize $ \color{lightgray} \pm 1.9$ } & -\\
                                            & NVS $\lambda$=0.5             & $34.4${\scriptsize$\color{lightgray} \pm0.2$} & $0.461${\scriptsize $\color{lightgray} \pm 0.004 $ }   & -                                                                 & -                                                             & $65.6${\scriptsize $  \color{lightgray} \pm 0.8$ }                        & $ 86.1${\scriptsize $ \color{lightgray} \pm 0.7 $  }                  & $25.5${\scriptsize $ \color{lightgray} \pm 0.2$}   & - & -\\
    \midrule
     \multirow{6}{*}{\rotatebox{90}{DE-EN}} & -                             & $40.7${\scriptsize$\color{lightgray} \pm0.2$} & $0.645${\scriptsize $ \color{lightgray} \pm 0.001$  }  & $\color{lightgray} 100$                                           & $\color{lightgray} 100$                                       & $ \color{lightgray} 1128  ${\scriptsize $\color{lightgray}   \pm 17$  }   & $ \color{lightgray} 229 ${\scriptsize $\color{lightgray} 	 \pm 1 $  } & $29.8${\scriptsize $  \color{lightgray} \pm 0.1$ } & $ \color{lightgray} 100 $                                    & $\color{lightgray} 100$  \\ 
                                            & align k=200                   & $40.7${\scriptsize$\color{lightgray} \pm0.2$} & $0.643${\scriptsize $ \color{lightgray} \pm 0.002$  }  & $\phantom{0}98.1 ${\scriptsize $\color{lightgray} \pm 2.6$  }     & $\phantom{0}98.8${\scriptsize $ \color{lightgray} \pm 2.3$  } & $53.6 ${\scriptsize $ \color{lightgray} \pm 0.5$  }                       & $ 76.9 ${\scriptsize $ \color{lightgray} \pm 0.7 $ }                  & $29.8 ${\scriptsize $ \color{lightgray} \pm 0.1$ } & $\phantom{0}96.9 ${\scriptsize $\color{lightgray} \pm 2.1$}  & $\phantom{0}97.9 ${\scriptsize $ \color{lightgray} \pm 2.2$} \\
                                            & align k=1000                  & $40.7${\scriptsize$\color{lightgray} \pm0.2$} & $0.645${\scriptsize $ \color{lightgray} \pm 0.001$  }  & $\phantom{0}99.3${\scriptsize $ \color{lightgray} \pm 2.5$ }      & $100.3${\scriptsize $ \color{lightgray} \pm 2.2$  }           & $67.9 ${\scriptsize $ \color{lightgray} \pm 0.5$  }                       & $ 84.4 ${\scriptsize $ \color{lightgray} \pm 0.6$  }                  & $29.8 ${\scriptsize $ \color{lightgray} \pm 0.1$ } & $101.2 ${\scriptsize $\color{lightgray} \pm 2.1$          }  & $101.8 ${\scriptsize $ \color{lightgray} \pm 2.2$} \\
                                            & NVS $\lambda$=0.99            & $40.7${\scriptsize$\color{lightgray} \pm0.1$} & $0.644${\scriptsize $ \color{lightgray} \pm 0.003$  }  & -                                                                 & $\phantom{0}99.8${\scriptsize $ \color{lightgray} \pm 2.4$  } & $47.6 ${\scriptsize $ \color{lightgray} \pm 0.6$  }                       & $ 76.4 ${\scriptsize $ \color{lightgray} \pm 0.7 $ }                  & $29.7 ${\scriptsize $ \color{lightgray} \pm 0.1$ } & -                                                            & $\phantom{0}99.9 ${\scriptsize $\color{lightgray} \pm 2.1$         }  \\
                                            & NVS $\lambda$=0.9             & $40.7${\scriptsize$\color{lightgray} \pm0.2$} & $0.645${\scriptsize $ \color{lightgray} \pm 0.002 $}   & $101.6 ${\scriptsize $ \color{lightgray} \pm 2.6$        }        & -                                                             & $54.4 ${\scriptsize $ \color{lightgray} \pm 0.6$  }                       & $ 79.8 ${\scriptsize $ \color{lightgray} \pm 1.3 $ }                  & $29.7 ${\scriptsize $ \color{lightgray} \pm 0.1$ } & $101.0 ${\scriptsize $\color{lightgray} \pm 2.2$          }  & - \\
                                            & NVS $\lambda$=0.5             & $40.7${\scriptsize$\color{lightgray} \pm0.1$} & $0.645${\scriptsize $ \color{lightgray} \pm 0.002 $}   & -                                                                 & -                                                             & $63.9 ${\scriptsize $ \color{lightgray} \pm 0.5$  }                       & $ 83.2 ${\scriptsize $ \color{lightgray} \pm 0.7 $ }                  & $29.7 ${\scriptsize $ \color{lightgray} \pm 0.1$ } & - & -\\
    \midrule
     \multirow{6}{*}{\rotatebox{90}{EN-RU}} & -                             & $23.6${\scriptsize$\color{lightgray} \pm0.1$} & $ 0.528${\scriptsize $ \color{lightgray} \pm 0.002 $ } & $\color{lightgray} 100$                                           & $\color{lightgray} 100$                                       & $\color{lightgray}  673 ${\scriptsize $\color{lightgray}  \pm 7$  }       & $ \color{lightgray} 145  ${\scriptsize $\color{lightgray}    \pm 1 $} \\ 
                                            & align k=200                   & $23.3${\scriptsize$\color{lightgray} \pm0.1$} & $ 0.507${\scriptsize $ \color{lightgray} \pm 0.003 $ } & $\phantom{0}95.9 ${\scriptsize $\color{lightgray} \pm 1.8$    }   & $\phantom{0}94.1${\scriptsize $ \color{lightgray} \pm 1.8$  } & $49.9 ${\scriptsize $  \color{lightgray} \pm 0.8$  }                      & $ 78.2 ${\scriptsize $ \color{lightgray} \pm 0.9 $} \\
                                            & align k=1000                  & $23.5${\scriptsize$\color{lightgray} \pm0.1$} & $ 0.527${\scriptsize $ \color{lightgray} \pm 0.003 $ } & $100.4 ${\scriptsize $ \color{lightgray} \pm 1.8$     }           & $\phantom{0}99.9${\scriptsize $ \color{lightgray} \pm 1.7$  } & $64.8 ${\scriptsize $  \color{lightgray} \pm 0.8$  }                      & $ 82.9 ${\scriptsize $ \color{lightgray} \pm 0.7 $} \\
                                            & NVS $\lambda$=0.99            & $23.6${\scriptsize$\color{lightgray} \pm0.1$} & $ 0.525${\scriptsize $ \color{lightgray} \pm 0.002 $ } & -                                                                 & $\phantom{0}99.2${\scriptsize $ \color{lightgray} \pm 1.7$  } & $48.1 ${\scriptsize $  \color{lightgray} \pm 0.6$  }                      & $ 79.8 ${\scriptsize $ \color{lightgray} \pm 0.9 $} \\
                                            & NVS $\lambda$=0.9             & $23.6${\scriptsize$\color{lightgray} \pm0.1$} & $ 0.528${\scriptsize $ \color{lightgray} \pm 0.003 $ } & $\phantom{0}99.3 ${\scriptsize $\color{lightgray} \pm 1.7$  }     & -                                                             & $58.2 ${\scriptsize $  \color{lightgray} \pm 0.8$  }                      & $ 82.2 ${\scriptsize $ \color{lightgray} \pm 0.9 $} \\
                                            & NVS $\lambda$=0.5             & $23.6${\scriptsize$\color{lightgray} \pm0.0$} & $ 0.529${\scriptsize $ \color{lightgray} \pm 0.002 $ } & -                                                                 & -                                                             & $67.5 ${\scriptsize $  \color{lightgray} \pm 0.8$  }                      & $ 86.1 ${\scriptsize $ \color{lightgray} \pm 0.7 $} \\
    \midrule
     \multirow{6}{*}{\rotatebox{90}{RU-EN}} & -                             & $35.5${\scriptsize$\color{lightgray} \pm0.1$} & $ 0.559${\scriptsize $ \color{lightgray} \pm 0.005 $}  & $\color{lightgray} 100$                                           & $\color{lightgray} 100$                                       & $\color{lightgray} 568 ${\scriptsize $\color{lightgray}  \pm 8 $        } & $ \color{lightgray}  123${\scriptsize  $\color{lightgray} \pm 1 $ } \\ 
                                            & align k=200                   & $35.2${\scriptsize$\color{lightgray} \pm0.1$} & $ 0.552${\scriptsize $ \color{lightgray} \pm 0.004 $}  & $\phantom{00}96.4 ${\scriptsize $ \color{lightgray} \pm 2.5$    } & $\phantom{0}97.7${\scriptsize $ \color{lightgray} \pm 2.5$  } & $49.3 ${\scriptsize $ \color{lightgray} \pm 0.9$ }                        & $ 79.4 ${\scriptsize  $\color{lightgray} \pm 1.0$ } \\
                                            & align k=1000                  & $35.4${\scriptsize$\color{lightgray} \pm0.1$} & $ 0.561${\scriptsize $ \color{lightgray} \pm 0.004 $}  & $\phantom{00}99.9 ${\scriptsize $ \color{lightgray} \pm 2.5$  }   & $\phantom{0}98.7${\scriptsize $ \color{lightgray} \pm 2.3$  } & $62.7 ${\scriptsize $ \color{lightgray} \pm 0.5$ }                        & $ 81.8 ${\scriptsize  $\color{lightgray} \pm 0.8 $ } \\
                                            & NVS $\lambda$=0.99            & $35.5${\scriptsize$\color{lightgray} \pm0.1$} & $ 0.558${\scriptsize $ \color{lightgray} \pm 0.004 $}  & -                                                                 & $100.2${\scriptsize $ \color{lightgray} \pm 2.5$  }           & $47.2 ${\scriptsize $ \color{lightgray} \pm 0.5$ }                        & $ 79.9 ${\scriptsize  $\color{lightgray} \pm 1.0$ } \\
                                            & NVS $\lambda$=0.9             & $35.5${\scriptsize$\color{lightgray} \pm0.1$} & $ 0.559${\scriptsize $ \color{lightgray} \pm 0.005 $}  & $\phantom{0}101.1 ${\scriptsize $ \color{lightgray} \pm 2.6$   }  & -                                                             & $52.2 ${\scriptsize $ \color{lightgray} \pm 0.6$ }                        & $ 80.9 ${\scriptsize  $\color{lightgray} \pm 1.3$ } \\
                                            & NVS $\lambda$=0.5             & $35.5${\scriptsize$\color{lightgray} \pm0.1$} & $ 0.559${\scriptsize $ \color{lightgray} \pm 0.005 $}  & -                                                                 & -                                                             & $60.2 ${\scriptsize $ \color{lightgray} \pm 0.6$ }                        & $ 83.6 ${\scriptsize  $\color{lightgray} \pm 0.9$ }\\

 \end{tabular}
  \caption{Experimental results for unconstrained decoding (baseline), alignment-based VS with different $k$, and Neural Vocabulary Selection with varying $\lambda$. BLEU and COMET: mean and std of three runs with different random seeds. Human evaluation: source-based direct assessment renormalized so that the unconstrained baseline is at 100\%, with 95\% CI of a paired t-test. We ran two sets of human evaluations comparing 4 systems each. CPU/GPU: p90 latency in ms with 95\% CI based on 30 runs with batch size 1, shown as a percentage of the baseline. }
  \label{tab:experimental-results}
\end{table*}

\section{Experiments}
\label{sec:experiments}

\subsection{Setup}

\label{exp_setup}

Our training setup is guided by best practices for efficient \gls*{nmt} to provide a strong low latency baseline: deep Transformer as encoder with a lightweight recurrent unit in the shallow decoder~\citep{berard2021efficient,kim-etal-2019-research}, int8 quantization for CPU and half-precision GPU inference.
We use the constrained data setting from WMT20 \citep{WMT20Findings} with four language pairs English-German, German-English, English-Russian, Russian-English and apply corpus cleaning heuristics based on sentence length and language identification.
We tokenize with \textit{sacremoses}\footnote{\url{https://github.com/alvations/sacremoses}} and byte-pair encode~\citep{sennrich-etal-2016-neural} the data  with 32k merge operations.

All models are Transformers~\citep{Vaswani:2017} trained with the Sockeye 2 toolkit~\citep{domhan2020sockeye}.
We release the \gls*{nvs} code as part of the Sockeye toolkit\footnote{\url{https://github.com/awslabs/sockeye/tree/naacl2022}}.
We use a 20-layer encoder and a 2-layer decoder with self-attention replaced by \glspl*{ssru}~\citep{kim-etal-2019-research}.

\gls*{nvs} and \gls*{nmt} objectives are optimized jointly, but gradients of the \gls*{nvs} objective are blocked before the encoder. This allows us to compare the different vocabulary selection techniques on the same translation model that is unaffected by the choice of vocabulary selection.
All vocabulary selection methods operate at the BPE level.
We use the translation dictionaries from \texttt{fast\_align} for alignment-based vocabulary selection.
We use a minimum of $k=200$ for alignment-based vocabulary selection which is at the upper end of what is found in previous work.
\citet{junczys-dowmunt-etal-2018-marian-cost} set $k=100$, \citet{kim-etal-2019-research} set $k=75$, and \citet{shi-knight-2017-speeding} set $k=50$.
Smaller $k$ would lead to stronger quality degradations at lower latency.
GPU and CPU latency is evaluated at single-sentence translation level to match real-time translation use cases where latency is critical.
We evaluate translation quality using SacreBLEU~\citep{post-sacrebleu}\footnote{BLEU+case.mixed+lang.en-de+numrefs.1+smooth.exp+tok.13a+version.1.4.14.} and COMET~\citep{rei-etal-2020-comet}\footnote{wmt-large-da-estimator-1719}.
Furthermore, we conduct human evaluations with two annotators on the subsets of newstest2020 and IDTS test sentences where outputs differ between NVS $\lambda=0.9$ (0.99) and align $k=200$.
Professional bilingual annotators rate outputs of four systems concurrently in absolute numbers with increments of 0.2 from 1~(worst) to 6~(best).
Ratings are normalized so that the (unconstrained) baseline is at 100\%. 
Complementary details on the training setup, vocabulary selection model size, human and latency evaluation setup can be found in Appendix~\ref{app:reproducibility}.

\subsection{Results}

Table~\ref{tab:experimental-results} shows results of different vocabulary selection models on newstest2020 and the ITDS idiom set, compared to an unconstrained baseline without vocabulary selection.
Automatic evaluation metrics show only very small differences between models.
For three out of four language pairs, the alignment model with $k=200$ performs slightly worse than the unconstrained baseline (0.2-0.3 BLEU). This corroborates existing work that quality measured by automatic metrics is not significantly affected by alignment-based vocabulary selection~\cite{jean-etal-2015-using,shi-knight-2017-speeding,kim-etal-2019-research}.

However, human-perceived quality of alignment-based vocabulary selection with $k=200$ is consistently lower than the baseline.
COMET, found to correlate better with human judgements than BLEU~\cite{kocmi2021ship}, only reflects this drop in two out of the four language pairs, considering confidence intervals across random seeds.
Increasing $k$ to 1000 closes the quality gap with respect to human ratings taking the confidence intervals into account.
The same is true for vocabulary selection using \gls*{nvs} at both $\lambda=0.9$ and $\lambda=0.99$, where quality is also within the confidence intervals of the unconstrained baseline.
However, \gls*{nvs} is consistently faster than the alignment-based model.
For $\lambda=0.9$ we see CPU latency improvements of 95 ms on average across language arcs.
Increasing the threshold to $\lambda=0.99$ latency compared to $k=1000$ is reduced by 157 ms on average.
The same trend holds for GPU latency but with smaller differences.
\begin{figure*}[th]
  \centering
  \includegraphics[width=\textwidth]{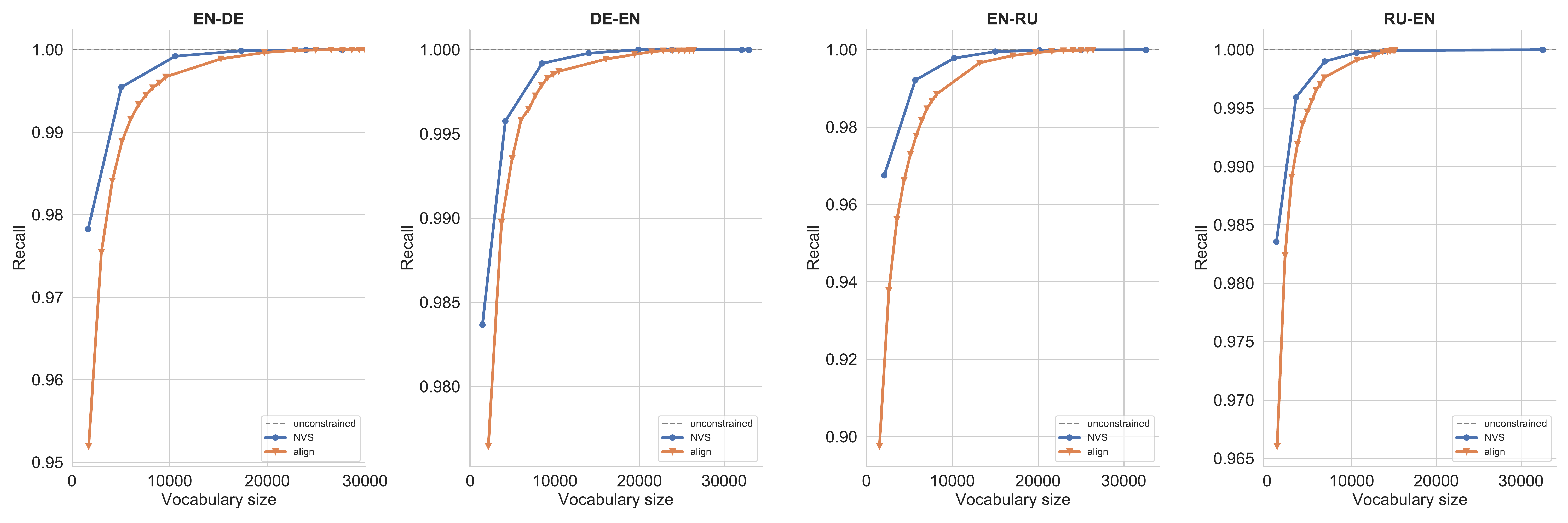}
  \caption{Vocabulary size (speed) vs. recall of reference tokens (quality) for newstest2020. For \gls*{nvs}, values correspond to $\lambda \in [0.99, 0.9, 0.5, 0.1, 0.01, .., 0.000001]$. For align, values correspond to $k \in [100, 200, .., 1000, 2000, .., 10000]$.}
  \label{fig:vsize_recall}
\end{figure*}
Figure~\ref{fig:vsize_recall} compares the \gls*{nvs} model against the alignment model according to the speed/quality tradeoff reflected by average vocabulary size vs. reference token recall on newstest2020.
\gls*{nvs} consistently outperforms the alignment model, especially for small average vocabulary sizes where \gls*{nvs} achieves substantially higher recall.
This demonstrates that the reduced vocabulary size and therefore faster decoder steps can amortize the cost of running the lightweight \gls*{nvs} model, which is fully parallelized across source tokens as part of the encoder.

\begin{figure*}[th]
  \centering
  \includegraphics[width=\textwidth]{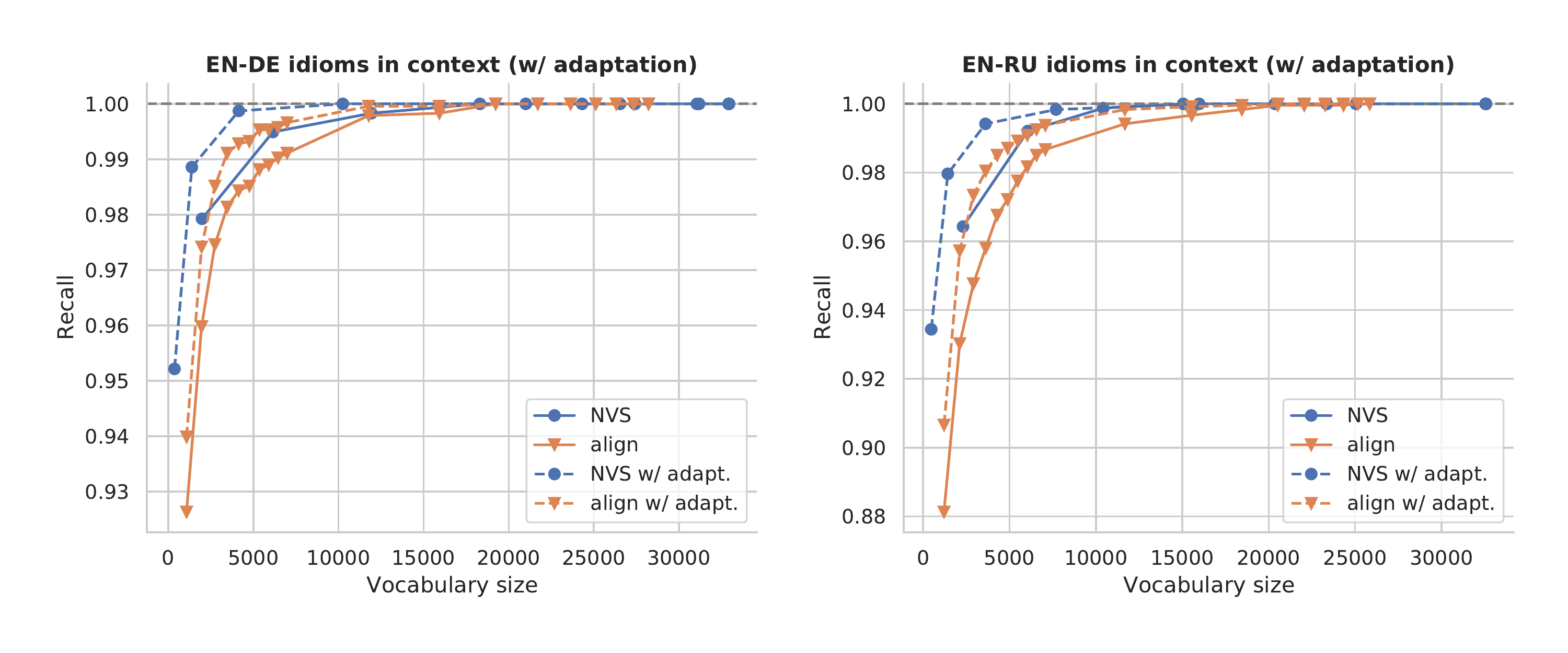}
  \caption{Vocabulary size (speed) vs recall of reference tokens (quality) for the internal Idioms test set \emph{with} and \emph{without} adapting on idioms for \gls*{nvs} and align models. For \gls*{nvs}, the values correspond to setting $\lambda \in [0.99, 0.9, 0.5, 0.1, 0.01, .., 0.000001]$. For align, the values correspond to setting $k \in [100, 200, .., 1000, 2000, .., 10000]$.}
  \label{fig:appendix:vsize_recall_idioms_adapt}
\end{figure*}

To evaluate a domain adaptation setting, we fine-tune the \gls*{nvs} models on a set of 300 held-out sentences of idioms in sentence context for 10 epochs.
For a fair comparison, we also include the same data for the alignment-based vocabulary selection.
Figure~\ref{fig:appendix:vsize_recall_idioms_adapt} shows that \gls*{nvs} yields pareto optimality over the alignment model with and without domain adaptation to a small internal training set of idiomatic expressions in context.
This highlights the advantage of \gls*{nvs} which is automatically updated during domain fine-tuning as it is part of a single model. 
See Appendix~\ref{appendix:plots} for additional figures on the proverbs and ITDS test sets, where the same trend holds.

\subsection{Analysis}
Our proposed neural vocabulary selection model benefits from contextual target word prediction.
We demonstrate this by comparing the predicted \gls*{bow} when using the source sentence context versus predicting \gls*{bow}s individually for each input word (which may consist of multiple subwords) and taking the union of individual bags.
We use the NVS models that are adapted to a set of idiomatic expressions for this analysis to ensure that the unconstrained baseline models produce reasonable translations for the Idiom test set.

\begin{table}[ht]
\centering
\small
\begin{tabular}{lllcc}
 & \bf Ref tokens & \bf $\lambda$ & \bf Context & \bf No Context \\
\midrule
\multirow{4}{*}{EN-DE} & \multirow{2}{*}{All} & 0.9 & 0.93 & 0.62 \\
& & 0.99 & 0.73 & 0.31 \\
& \multirow{2}{*}{All excl} & 0.9 & 0.32 & 0.01 \\
& & 0.99 & 0.43 & 0.01 \\
\midrule
\multirow{4}{*}{EN-RU} & \multirow{2}{*}{All} & 0.9 & 0.75 & 0.43 \\
&  & 0.99 & 0.57 & 0.25 \\
& \multirow{2}{*}{All excl} & 0.9 & 0.37 & 0.05 \\
&  & 0.99 & 0.38 & 0.06 \\
\end{tabular}
\caption{Percentage of segments with all idiomatic reference tokens included in the \gls*{bow} (All), or exclusively included in the contextual or non-contextual \gls*{bow} (All excl) for \gls*{nvs} threshold $\lambda$.}
\label{tab:context_vs_nocontext}
\end{table}

Table~\ref{tab:context_vs_nocontext} shows the percentage of segments for which all reference tokens are included in the contextual vs. the non-contextual \gls*{bow} for an acceptance threshold of 0.9 and 0.99. Independent of the threshold, predicting the \gls*{bow} using source context yields significantly larger overlap with idiomatic reference tokens. 
We also measure the extent to which idiomatic reference tokens are included exclusively in the contextual or non-contextual \gls*{bow}. For 32\% of EN-DE segments, only the contextual \gls*{bow} contains all idiomatic reference tokens. For non-contextual \gls*{bow}s, this happens in only 1\% of the segments (with $\lambda$=0.9). For EN-RU, the values are 38\% versus 6\%, respectively. This shows that the model makes extensive use of contextualized source representations in predicting the relevant output tokens for idiomatic expressions.

Figure~\ref{fig:context_examples} shows a few illustrative examples where the idiomatic reference is only reachable with the contextual \gls*{bow} prediction. Consider the last example containing the English idiom ``to wrap one's head around it''. Even though the phrase is rather common in English, the German translation ``verstehen'' (\textit{to understand}) would not be expected to rank high for any of the idiom source tokens. Evaluating the tokens in context however yields the correct prediction.

\begin{figure*}[ht]
\centering
\small
\renewcommand{\arraystretch}{1.2}
\begin{tabular}{@{} p{11.5cm} p{3.5cm} @{}}
\bf Source & \bf Idiomatic target \\
\midrule
I thought I would be nervous , but I was \textbf{cool as a cu\pmboxdrawuni{2581} cum\pmboxdrawuni{2581} ber} . & die \textbf{Ruhe} selbst \\
He decides that it is better \textbf{to face the music} , op\pmboxdrawuni{2581} ting to stay and conf\pmboxdrawuni{2581} ess . & sich den \textbf{Dingen} stellen \\
The Classic Car Show is held in conjunction with Old Sett\pmboxdrawuni{2581} ler 's Day , \textbf{rain or shine} . & bei \textbf{jedem} Wetter \\
Tools , discipline , formal methods , process , and profession\pmboxdrawuni{2581} alism were tou\pmboxdrawuni{2581} ted as \textbf{silver bul\pmboxdrawuni{2581} lets} : & \textbf{Wunder\pmboxdrawuni{2581}} wa\pmboxdrawuni{2581} ffe \\
They said he was ' a little bit \textbf{under the weather} ' . & sich nicht \textbf{wohl\pmboxdrawuni{2581} fühlen} \\
I still can 't \textbf{wra\pmboxdrawuni{2581} p my head around it} . & \textbf{verstehen} \\
\end{tabular}
\caption{Test inputs from the internal Idioms test set for which the highlighted tokens in the idiomatic reference are \textit{exclusively} included in the contextual BOW (computed for idiom-adapted \gls*{nvs} model with $\lambda=0.9$).}
\label{fig:context_examples}
\end{figure*}

\section{Related work}
There are two dominant approaches to generate a restricted set of target word candidates (i) using an external model and (ii) using the \gls*{nmt} system itself.

In the first approach, a short-list of translation candidates is generated from word-alignments \cite{jean-etal-2015-using,kim-etal-2019-research}, phrase table, and the most common target words  \cite{mi-etal-2016-vocabulary}. \citet{vocabularySelectionStrategies} propose an additional method using support vector machines to predict target candidates from a sparse representation of the source sentence.

In the second approach, \citet{sankaran2017attention} build alignment probability table from the soft-attention layer from decoder to encoder. However, applying their method to multi-head attention in Transformer is non-trivial as attention may not capture word-alignments in multiple attention layers~\cite{li-etal-2019-word}. \citet{shi-knight-2017-speeding}~use local sensitive hashing to shrink the target vocabulary during decoding, though their approach only reduces latency on CPUs instead of GPUs. 

\citet{chen2019learning} reduce the softmax computation by first predicting a cluster of target words and then perform exact search (i.e., softmax) on that cluster. The clustering process is trained jointly with the translation process in their approach.

Closely related to our work is \citet{weng-etal-2017-neural}, who predict all words in a target sentence from the initial hidden state of the decoder.
Our NVS model differs from theirs in that we make a prediction for each source token and aggregate the results via max-pooling to scale with sentence length.
Recent work of~\citet{bogoychev-chen-2021-highs} illustrates the risk associated with reducing latency via vocabulary selection in domain-mismatched settings.
Our work takes this a step further by providing a detailed analysis on the shortcomings of vocabulary selection and proposing a model to mitigate them. %

Related to our findings on non-compositional expressions, \citet{renduchintala-etal-2021-gender} evaluate the effect of methods used to speed up decoding in Transformer models on gender bias and find minimal BLEU degradations but reduced gendered noun translation performance on a targeted test set.

\section{Conclusions}
Alignment-based vocabulary selection is a common method to heavily constrain the set of allowed output words in decoding for reduced latency with only minor BLEU degradations.
We showed with human evaluations and a targeted qualitative analysis that such translations are perceivably worse.
Even recent automatic metrics based on pre-trained neural networks, such as COMET, are only able to capture the observed quality degradations in two out of four language pairs.
Human-perceived quality is negatively affected both for generic translations, represented by newstest2020, as well as for idiomatic translations.
Increasing the vocabulary selection threshold can alleviate the quality issues at an increased single sentence translation latency.
To preserve both translation latency and quality we proposed a neural vocabulary selection model that is directly integrated into the translation model.
Such a joint model further simplifies the training pipeline, removing the dependency on a separate alignment model.
Our model has higher reference token recall at similar vocabulary sizes, translating into higher quality at similar latency.

\bibliography{anthology,custom}
\bibliographystyle{acl_natbib}

\clearpage
\appendix

\section{Reproducibility Details}
\label{app:reproducibility}

\paragraph{Data}
We use the constrained data setting from WMT20 \citep{WMT20Findings} with four language pairs English-German, German-English, English-Russian, Russian-English.
Noisy sentence pairs are removed based on heuristics, namely sentences with a length ratio $>1.5$, $> 70\%$ token overlap, $> 100$ BPE tokens and those where source or target language does not match according to LangID~\citep{LangID} are filtered.

\paragraph{Model} We train pre-norm Transformer~\citep{Vaswani:2017} models
with an embedding dimension of 1024 and a hidden dimension of 4096.

\begin{table}[th]
\centering
 \begin{tabular}{l r} 
 \bf Model & \bf Size \\
 \midrule
 align k=200 & 6,590,600 \\
 align k=1000 & 32,953,000 \\
 NVS & 33,776,825 \\
 \end{tabular}
 
 \caption{Model size in terms of in-memory float numbers for EN-DE model with a target vocabulary size of 32953. This does not reflect actual memory consumption, as the computation of the NVS layer may require more intermediate memory.}
 \label{tab:modelsize}
\end{table}

Table~\ref{tab:modelsize} compares the memory consumption of the different vocabulary selection models in terms of float numbers. We see that the \gls*{nvs} model requires a similar number of floating point numbers as the alignment-based model at $k=1000$.
Note, that this only represent the disk space requirements as other intermediate outputs would be required at runtime for either vocabulary selection model.

\paragraph{Training} The NMT objective uses label smoothing with constant 0.1, the \gls*{nvs} objective sets the positive class weight $\lambda_p$ to 100,000. Models train on 8 Nvidia Tesla V100 GPUs on AWS p3.16xlarge instances with an effective batch size of 50,000 target tokens accumulated over 40 batches. We train for 70k updates with the Adam~\citep{kingma2014adam} optimizer, using an initial learning rate of 0.06325 and linear warmup over 4000 steps. Checkpoints are saved every 500 updates and we average the weights of the 8 best checkpoints according to validation perplexity.

\paragraph{Inference} For GPU latency, we run in half-precision mode (FP16) on AWS g4dn.xlarge instances. CPU benchmarks are run with INT8 quantized models run on AWS c5.2xlarge instances.
We decode using beam search with a beam of size 5.
Each test set is decoded 30 times on different hosts, and we report the mean p90 latency with its 95\% confidence interval. Alignment-based vocabulary selection includes the top $k$ most frequently aligned BPE tokens for each source token based on a \texttt{fast\_align} model trained on the same data as the translation model. \gls*{nvs} includes all tokens that are scored above the threshold $\lambda$. All vocabulary selection methods operate at the BPE level.

\paragraph{Evaluation}
Human Evaluations and COMET / BLEU use full precision (FP32) inference outputs. 
We decided to use FP32 for human evaluation as we wanted to evaluate the quality of the underlying model independent of whether it gets used on CPU or GPU and the output differences between FP16/FP32/INT8 being small.
We report mean and standard deviation of SacreBLEU~\citep{post-sacrebleu}\footnote{BLEU+case.mixed+lang.en-de+numrefs.1+smooth.exp+tok.13a+version.1.4.14.} and COMET~\citep{rei-etal-2020-comet} scores on detokenized outputs for three runs with different random seeds.
For human evaluations, bilingual annotators see a source segment and the output of a set of 4 systems at once when assigning an absolute score to each output.
The size of the evaluation set was 350 for EN-DE and EN-RU and 200 for DE-EN and RU-EN for newstest2020. We used the full sets of sentences differing between NVS $\lambda=0.9$, align $k=200$ for the ITDS test set (309 for EN-DE and 273 for DE-EN).

\paragraph{Adaptation}
For domain adaptation, we fine-tune the \gls*{nvs} model for 10 epochs using a learning rate of $0.0001$ and a batch size of 2048 target tokens.
To adapt the alignment-based vocabulary selection model, we include the adaptation data as part of the training data for the alignment model. We upsample the adaptation data by a factor of 10 for a comparable setting with \gls*{nvs} fine-tuning.

\section{Positive class weight ablation}
\label{appendix:posweight}
\begin{table}[t!]
\centering
\begin{tabular}{llcc}

 & \textbf{pos. weight} & \textbf{BLEU} & \textbf{COMET} \\
EN-DE  & auto $x=10$ & 34.4 & 0.459 \\
       & auto $x=1$  & 34.1 & 0.458 \\
       & 100k     & 34.4 & 0.461 \\
       & 10k      & 34.2 & 0.460 \\
       & 1k       & 34.4 & 0.463 \\
       & 100      & 34.2 & 0.456 \\
       & 10       & 32.5 & 0.295 \\
       & 1        & 15.9 & -0.498 \\
 \midrule
 DE-EN & auto $x=10$ & 40.9 & 0.644 \\
       & auto $x=1$  & 40.8 & 0.640 \\
       & 100k     & 40.8 & 0.642 \\
       & 10k      & 41.0 & 0.645 \\
       & 1k       & 40.8 & 0.643 \\
       & 100      & 40.8 & 0.638 \\
       & 10       & 40.2 & 0.558 \\
       & 1        & 25.3 & -0.608 \\
 \midrule
EN-RU  & auto $x=10$ & 23.6 & 0.524 \\
       & auto $x=1$  & 23.5 & 0.524 \\
       & 100k     & 23.6 & 0.524 \\
       & 10k      & 23.7 & 0.528 \\
       & 1k       & 23.6 & 0.526 \\
       & 100      & 23.3 & 0.497 \\
       & 10       & 20.6 & 0.128 \\
       & 1        & 5.6  & -1.509 \\
 \midrule
RU-EN  & auto $x=10$ & 35.6 & 0.564 \\
       & auto $x=1$  & 35.6 & 0.563 \\
       & 100k     & 35.6 & 0.557 \\
       & 10k      & 35.8 & 0.565 \\
       & 1k       & 35.4 & 0.556 \\
       & 100      & 35.5 & 0.551 \\
       & 10       & 34.2 & 0.452 \\
       & 1  & 20.1  & -0.622 \\

\end{tabular}
\caption{Translation quality in terms of BLEU and COMET on newstest2020 with different weights for the positive class. \textit{auto $x$} refers to setting the weight according to the ratio of the negative class to the positive class with a factor $x$.}
\label{tab:posweight}
\end{table}

Based on preliminary experiments we had used a weight for the positive class ($\lambda_p$) of 100k in the experiments in \S\ref{sec:experiments}.
Here the positive class refers to tokens being present on the target side and the negative class to tokens being absent from the target side.
For a Machine Translation setting there are many more words that are not present than are present on the target side.
The negative class therefore dominates the positive class.
This can be counteracted by using a large value for the positive weight $\lambda_p$.

Instead of setting $\lambda_p$ to a fixed weight one can also define it as 
\begin{equation*}
  \lambda_p = x \frac{n_n}{n_p}
\end{equation*}
with $n_p$ as the number of unique target words, $n_n = V - n_p$ as the number of remaining words and $x$ being a factor to increase the bias towards recall.
This way the positive class and negative class are weighted equally.
Table~\ref{tab:posweight} shows the result of different positive weights, including the automatic setting according to the ratio (\textit{auto}).
We see that not increasing the weight of the positive class results in large quality drops.
For positive weights $>1000$ the quality differences are small.
The \textit{auto} setting provides an alternative that is easier to set than finding a fixed positive weight.

\section{Additional vocabulary size vs. recall plots}
\label{appendix:plots}
\begin{figure*}[th]
  \centering
  \includegraphics[width=\textwidth]{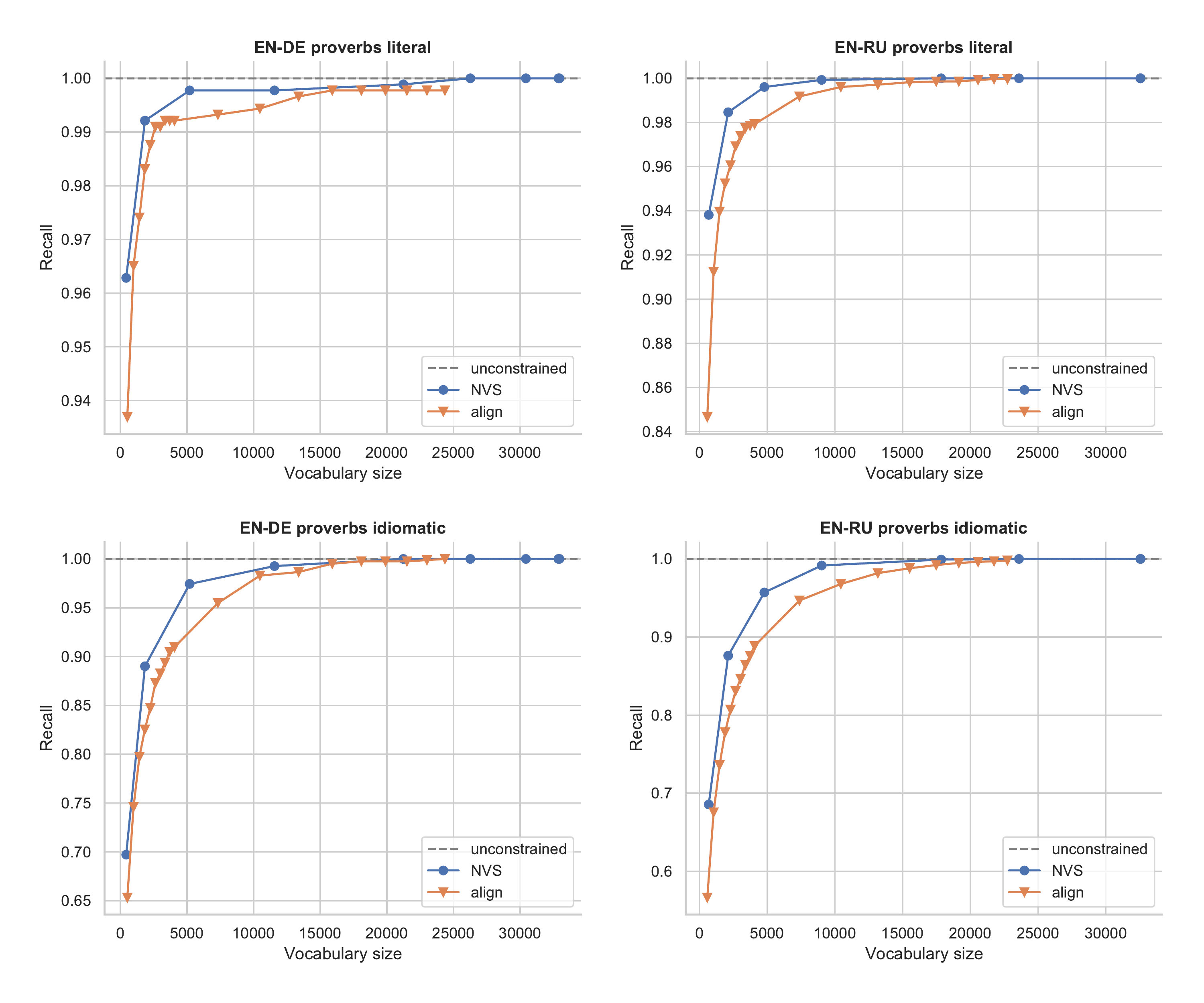}
  \caption{Vocabulary size (speed) vs. recall of reference tokens (quality) for proverbs test set. For \gls*{nvs}, the values correspond to setting $\lambda \in [0.99, 0.9, 0.5, 0.1, 0.01, .., 0.000001]$. For align, the values correspond to setting $k \in [100, 200, .., 1000, 2000, .., 10000]$.}
  \label{fig:appendix:vsize_recall_proverbs}
\end{figure*}

\begin{figure*}[th]
  \centering
\includegraphics[width=\textwidth]{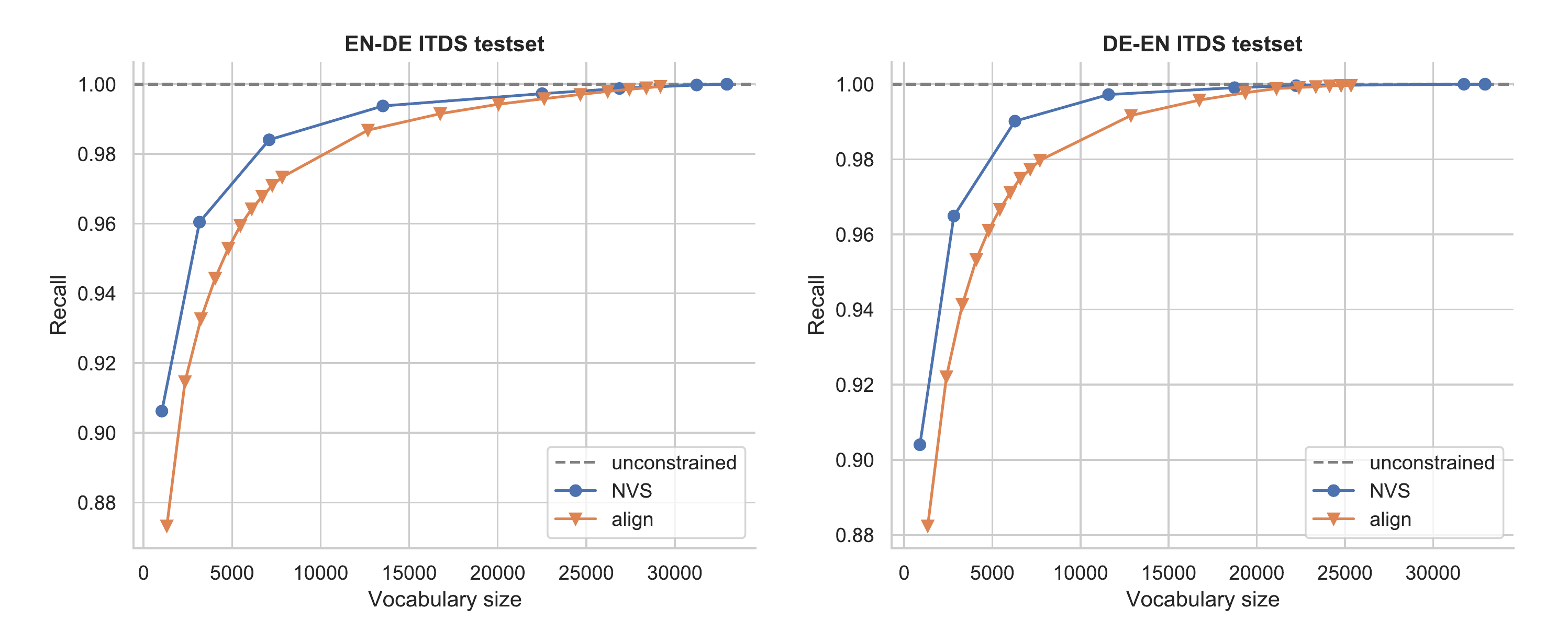}
  \caption{Vocabulary size (speed) vs. recall of reference tokens (quality) for ITDS test set. For \gls*{nvs}, the values correspond to setting $\lambda \in [0.99, 0.9, 0.5, 0.1, 0.01, .., 0.000001]$. For align, the values correspond to setting $k \in [100, 200, .., 1000, 2000, .., 10000]$.}
  \label{fig:appendix:vsize_recall_itds}
\end{figure*}

Figures~\ref{fig:appendix:vsize_recall_proverbs} and \ref{fig:appendix:vsize_recall_itds} provide results for the proverbs and ITDS test sets, respectively.
We see the same trend across all test sets of \gls*{nvs} offering higher recall at the same vocabulary size compared to alignment-based vocabulary selection.
For the proverbs test set this is true both for the literal and the idiomatic translations.

\end{document}